\documentclass{article}
\usepackage{spconf,amsmath,graphicx}
\usepackage{multirow}
\usepackage{multicol}
\usepackage{booktabs}
\usepackage{threeparttable}
\usepackage{arydshln}
\usepackage{stfloats}
\usepackage{supertabular}
\usepackage{subfigure}
\usepackage{bbding}
\usepackage{cite}

\title{Joint Multiple Intent Detection and Slot Filling via Self-distillation}
\name{Lisong Chen$^{1}$ \qquad Peilin Zhou$^1$ \qquad Yuexian Zou$^{1,2,*}$ 
}
\address{$^1$ADSPLAB, School of ECE, Peking University, Shenzhen, China\\$^2$Peng Cheng Laboratory, Shenzhen, China}
\begin{document}
%
\maketitle
\begin{abstract}
Intent detection and slot filling are two main tasks in natural language understanding (NLU) for identifying users' needs from their utterances. These two tasks are highly related and often trained jointly. However, most previous works assume that each utterance only corresponds to one intent, ignoring the fact that a user utterance in many cases could include multiple intents. In this paper, we propose a novel Self-Distillation Joint NLU model (SDJN) for multi-intent NLU. First, we formulate multiple intent detection as a weakly supervised problem and approach with multiple instance learning (MIL). Then, we design an auxiliary loop via self-distillation with three orderly arranged decoders: Initial Slot Decoder, MIL Intent Decoder, and Final Slot Decoder. The output of each decoder will serve as auxiliary information for the next decoder. With the auxiliary knowledge provided by the MIL Intent Decoder, we set Final Slot Decoder as the teacher model that imparts knowledge back to Initial Slot Decoder to complete the loop. The auxiliary loop enables intents and slots to guide mutually in-depth and further boost the overall NLU performance. Experimental results on two public multi-intent datasets indicate that our model achieves strong performance compared to others.

\end{abstract}
\section{Introduction}
\label{sec:intro}

\begin{figure}[t]
  \centering
  \includegraphics[width=8.cm]{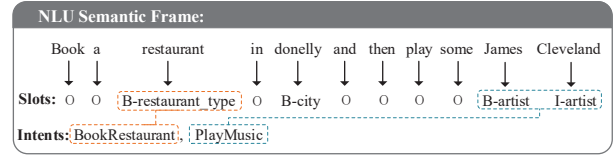}
  \caption{NLU semantic frame.}
  \label{fig:NLU_frame}
\end{figure}

In dialogue systems, the natural language understanding (NLU) component plays an important role. It consists of two sub-tasks, including intent detection and slot filling \cite{2011Spoken} which allow the dialogue system to create a semantic frame that summarizes the user's requests. As shown in Figure \ref{fig:NLU_frame}, intent detection is a classification task while slot filling is a sequence labeling task. Several text classification methods \cite{cortes1995support, bishop2006pattern, sarikaya2011deep} have been proposed to tackle intent detection. For slot filling, many sequence labeling methods \cite{berger1996maximum,toutanova2000enriching, mccallum2000maximum, lafferty2001conditional,jeong2008triangular, xu2013convolutional, yao2014spoken} have been explored. Taking a deeper look at the example, intent ``\textit{BookRestaurant}" is highly related to slot ``\textit{B$-$restaurant\underline{\hbox to 0.1cm{}}type}". This observation inspires many works \cite{liu2016attentionbased,goo2018slot,li2018self,Haihong2019ANB,Qin2019ASF,Zhou2020PINAN, huang2021sentiment} to jointly model the intent detection and slot filling, where the correlation between the intent and slots are utilized.

The works above mainly focus on the scenario where each utterance has only one intent. However, in many real-life situations, users may express multiple intents in an utterance, thus making it difficult to directly apply single intent NLU models. Therefore, great efforts are needed to develop NLU models that could handle such multi-intent problems. For multi-intent NLU, there are two main challenges: 1) correctly identifying multiple intents from a single utterance, especially when the intents are similar. 2) effectively enabling multiple intents to guide the corresponding token for slot prediction. \cite{gangadharaiah-narayanaswamy-2019-joint,qin-etal-2020-agif} are two works that focus on the multi-task framework for the multi-intent NLU. \cite{gangadharaiah-narayanaswamy-2019-joint} proposed an attention-based neural network that treats multi-intent detection as a multi-label classification task. Nevertheless, their model simply treats multiple intents information as an intent context vector and then incorporates it to guide slot directly. It failed to consider that the slots in an utterance may relate to different intents, just as shown in Figure \ref{fig:NLU_frame}. Based on this insight, \cite{qin-etal-2020-agif} proposed an intent-slot graph interaction layer to capture the interaction between multiple intents and each token.  

Though achieving promising performance for multi-intent NLU, their models still suffer some issues. For one thing, they adopt utterance context vector for multiple intent detection. Such utterance-level representations may miss out on fine-grained information that could be crucial to distinguishing intents. For another thing, they only consider the unidirectional interaction, namely using  intent to guide slot prediction, while slot can also offer important information for intent prediction.

In this paper, we propose a Self-distillation Joint NLU model (SDJN) to address the problems. First, we observe that intent information is mainly reflected from certain tokens. However, there are only utterance-level supervised labels available, so we formulate multiple intent detection as a weakly supervised problem and approach it with multiple instance learning (MIL) \cite{Keeler1991ASI}. In our case, we consider the tokens in the utterance as instances in MIL and predict their intent distributions to provide fine-grained intent information. The whole utterance is regarded as a bag, and the overall intent labels are the combination of instance predictions. Using the MIL method for intent detection is similar to Stack-Propagation framework in \cite{Qin2019ASF} where token-level intent prediction is performed. However, Stack-Propagation framework uses a simple voting mechanism to predict final intent which is only efficient for single intent detection. Instead of voting, we adopt an attention-based aggregation function which is effective under multiple intents scenario to predict overall intents. Second, inspired by self-distillation network \cite{Zhang2019BeYO}, we proposed a self-distillation method for joint NLU modeling by taking advantage of multi-task. In \cite{Zhang2019BeYO}, they propose a self-distillation network based on the depth of the network. Instead of deepening the network, the self-distillation method we proposed utilizes the multi-task information. The main idea behind our approach is to form an auxiliary loop. There are three decoders in SDJN, including Initial Slot Decoder, MIL Intent Decoder, and Final Slot Decoder arranged in order. The output of one decoder will serve as auxiliary information for the next decoder. With the intent information provided by MIL Intent Decoder, Final Slot Decoder generates better slot hidden states compare to Initial Slot Decoder. Therefore, we consider Final Slot Decoder as the teacher model that imparts knowledge back to Initial Slot Decoder to complete the auxiliary loop. It allows Initial Slot Decoder to learn a better intermediate representation from Final Slot Decoder. The workflow further establishes the interrelated connections for the multiple intents and slot information.

To summarize, the contributions of this paper are: (1) We formulate multiple intent detection as a weakly supervised problem and approach it with multiple instance learning.
(2) An improved self-distillation approach is proposed for improving joint modeling, allowing the model to form an auxiliary loop and exploits the interrelated connection between intent and slot information in depth. 
(3) Experimental results show that our approach achieve strong performance on two public multi-intent datasets (i.e., MixATIS and MixSNIPS~\cite{qin-etal-2020-agif}).

\begin{figure*}[tp]
    \subfigure[]{\includegraphics[width=0.7\textwidth]{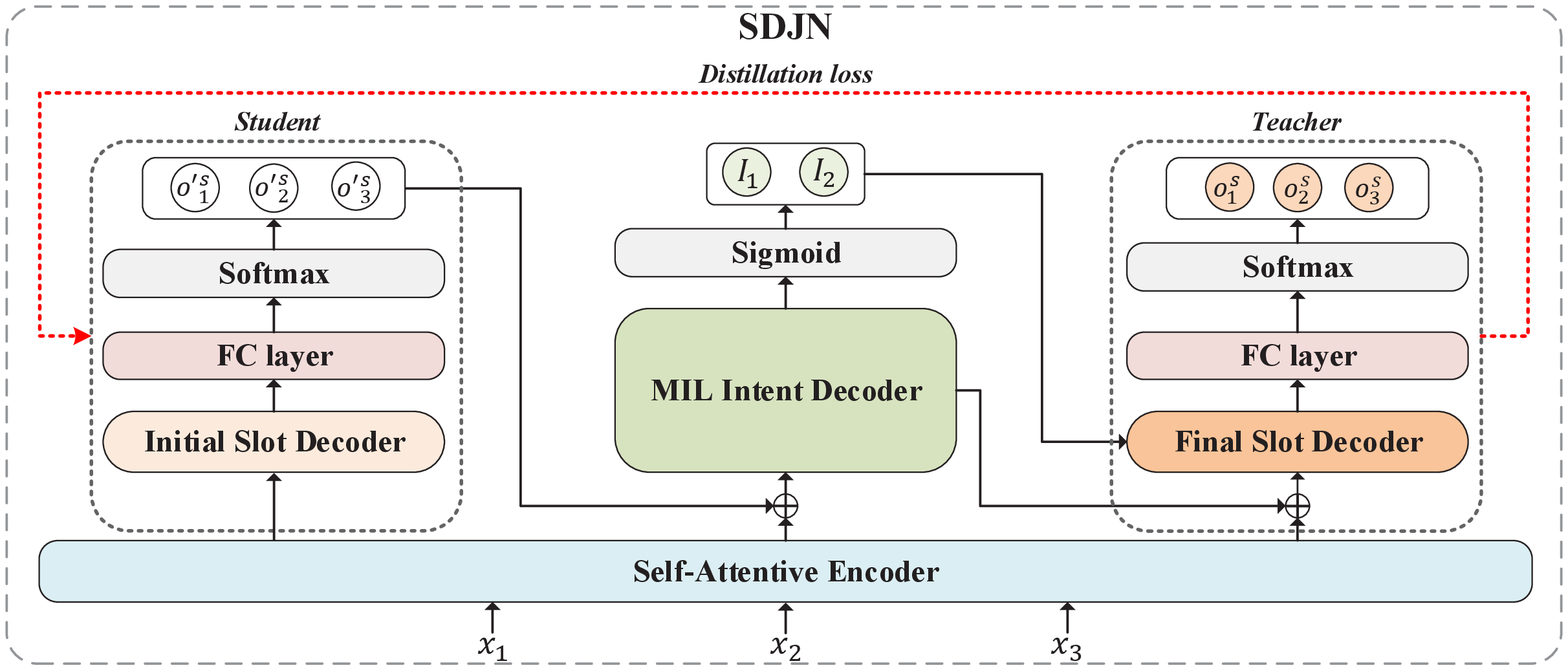}
    \label{fig:sdjn} }
     \subfigure[]{\includegraphics[width=0.18\textwidth]{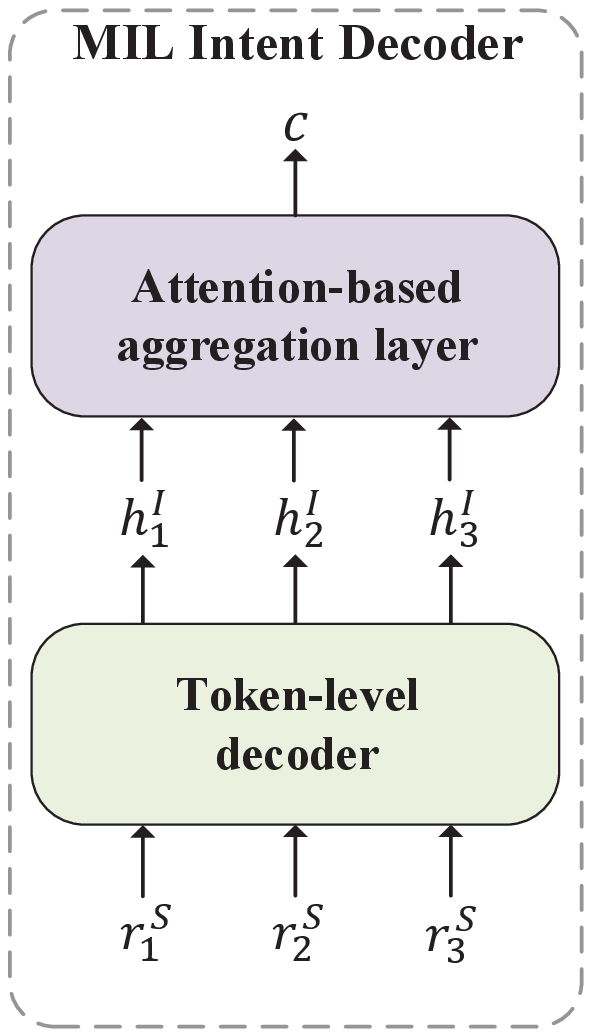}
    \label{fig:mil_intent} }
    \caption{The architecture of (a) SDJN model and (b) the MIL Intent Decoder}
    \label{fig:model_arch} 
\end{figure*}
 
\section{Approach}
\label{sec:typestyle}
In this section, we introduce our SDJN model in detail. The architecture of the model is illustrated in Figure \ref{fig:model_arch}. SDJN model consists of a shared encoder, three decoders, and a self-distillation process.

\subsection{Self-attentive Encoder}
\label{ssec:Self-attentive-Encoder}

Following the self-attentive encoder in ~\cite{Qin2019ASF,qin-etal-2020-agif}, we use a bidirectional LSTM (BiLSTM) \cite{1997Long} with self-attention mechanism \cite{2017Attention} as encoder to model temporal and contextual information from the utterance. The BiLSTM generates a series of context-sensitive hidden states $H$. Self-attention is expressive for both local and long-range dependencies which outputs a context-aware feature $A$. The final encoding representation $E$ is the concatenation of the outputs from BiLSTM and the self-attention mechanism:
\begin{equation}
    E = H \oplus A
\label{eq:encoder_out}
\end{equation}

\subsection{Initial Slot Decoder}
\label{ssec:encode}
In Initial Slot Decoder, it is going to decode initial slot fillings that will be used for guiding intents. We use a unidirectional GRU \cite{Cho2014Learning} for Initial Slot Decoder. The input feature is $E=\{e_1, ..., e_n\}$. At every decoding step $t$, the decoder state $h'^S_t$ can be formalized as:
\begin{equation}
    h'^S_t = f(h'^S_{t-1}, y'^S_{t-1}, e_t)
\label{eq:slot_dec}
\end{equation}
where $h'^S_{t-1}$ is the previous decoder state, $y'^S_{t-1}$ is the previous emitted slot prediction and  $e_t$ is the aligned encoder hidden state. The decoder state $h'^S_t$ will further be utilized to generate initial slot filling:
\begin{equation}
    y'^S_t = softmax(w_hh'^S_t)\\
\label{eq:slot_distribution}
\end{equation}
\begin{equation}
    o'^S_t = argmax(y'^S_t)
\label{eq:initial_slot}
\end{equation}
where $w_h$ are trainable parameters of the model and $O'^S=\{o'^S_1, ..., o'^s_n\}$ is the initial slot filling of the the utterance.  

\subsection{MIL Intent Decoder}
In our model, we approach multiple intent detection with MIL. Under MIL, the input utterance $X=\{x_1, ..., x_n\}$ is regarded as bag and the goal is to map each instance which is token $x_t$ in our case to intent label. The overall intents of the utterance will be the combination of token intents. 

As shown in Figure \ref{fig:sdjn}, we concatenate encoding representation $E$ and the initial slot information $O'^S$ to form the slot reinforce representation $R^S=\{r^S_1, ..., r^S_n\}$ as the input of MIL Intent Decoder. The MIL Intent Decoder consists of two components, as shown in Figure \ref{fig:mil_intent}: a token-level decoder and an aggregation layer. We use a unidirectional GRU as the token-level decoder. With the decoder state $h^I_t$ from token-level decoder, we use an attention-based prediction weighting module as aggregation layer:
\begin{equation}
    w_t=softmax(w_d h^I_t+b)
\label{eq:int_att}
\end{equation}
\begin{equation}
    c=\sum_tw_th^I_t
\label{eq:int_c}
\end{equation}
The $w_d$ is the trainable parameters and $w_t$ is the weight for each token. Through the aggregation layer, it rewards the tokens that provide more meaningful intent information. The overall intent distribution is calculated by $c$, the weighted sum of hidden representation $H^I=\{h^I_1, ..., h^I_n\}$, with a sigmoid activation:
\begin{equation}
    o^I=sigmoid(c)
\label{eq:int_label}
\end{equation}
$o^I=\{o^I_1, ..., o^I_{N_I}\}$ is the overall intent output of the utterance and $N_I$ is the amount of single intent labels. Since it is a multiple intent detection task, we apply a threshold $0<t_I<1.0$ to obtain intents. The predicted multiple intents $I=\{I_1, ..., I_m\}$ are those which probability are greater than $t_I$.

\subsection{Final Slot Decoder}
Final Slot Decoder is composed of two modules. A vanilla slot decoder that is identical to Initial Slot Decoder. And following \cite{qin-etal-2020-agif}, we incorporate the adaptive intent-slot graph interaction layer. First, we concatenate the encoding representation $E$ with hidden representation $H^I$ from MIL Intent Decoder to form intent reinforce representation $R^I=\{r^I_1, ..., r^I_n\}$. The vanilla slot decoder adopts $r^I_t$ to generate decoder state $h^S_t$. Instead of predicting the slot label with $h^S_t$ directly, it goes through the graph interaction layer. The graph interaction layer adopts the graph attention network (GAT) \cite{Velickovic2018GraphAN} to model the interrelation of intents and slots at the token level. Specifically, the slot hidden state $h^S_t$ from vanilla slot decoder and predicted multiple intents $I=\{I_1, ..., I_m\}$ are used as the initialized representation at $t$ time step $\Tilde{H}^{[0,t]}=\{h^S_t, \phi^{emb}(I_1), ..., \phi^{emb}(I_m)\}$ where $\phi^{emb}(\cdot)$ represents the embedding matrix of intents. Within the graph, the slot node representation in the $l$-th layer is calculated as:
\begin{equation}
    \Tilde{h}^{[l, t]}_i = \sigma(\sum\limits_{j\in \mathcal{N}_i}\alpha^{[l, t]}_{ij}W^{[l]}_h\Tilde{h}^{[l-1, t]}_j)
\label{eq:graph}
\end{equation}
$\Tilde{h}^{[l, t]}_i$ can be understood as node $i$ in the $l$-th layer of the graph which consists of the decoder state node and predicted intent nodes at $t$ time step. $\mathcal{N}_i$ is the first-order neighbors node $i$, $W_h$ is the trainable weight matrix, $\alpha_{ij}$ is the normalized attention weight and $\sigma$ represents the nonlinearity activation function.

Through $L$-layer of adaptive intent-slot graph interaction, we adopt the final slot hidden state representation $\Tilde{h}^{[L, t]}_0$ at $t$ time step for slot filling:
\begin{equation}
    y^S_t=softmax(W_s \Tilde{h}^{[L, t]}_0)
\label{eq:final_slot_dis}
\end{equation}
\begin{equation}
    o^S_t=argmax(y^S_t)
\label{eq:final_slot}
\end{equation}
where $o^S_t$ is the final predicted slot label of the $t$-th word in the utterance.

\subsection{Self Distillation and Joint Training}
In SDJN model, we propose a knowledge distillation method within a joint training model by taking advantage of multi-task. The teacher model is Final Slot Decoder while the student model is Initial Slot Decoder. We select Final Slot Decoder as the teacher model for the following reasons. On the input wise, Final Slot Decoder incorporates the token-level intent information to form intent reinforce representation for better decoding. On the structure-wise, Final Slot Decoder has an adaptive intent-slot graph interaction layer to correlate intent information with slots explicitly. Therefore, Final Slot Decoder is able to generate better output. To perform the distillation method, as illustrated in Figure \ref{fig:model_arch}, Final Slot Decoder provides the hint for Initial Slot Decoder. A hint is defined as the output of the hidden layers from the teacher model, whose aim is to guide the student model \cite{Romero2015FitNetsHF}. Specifically, we leverage the hidden state from Initial Slot Decoder and Final Slot Decoder to calculate the representative distance. The relation is obtained through the computation of the MSE loss. The implicit knowledge in Final Slot Decoder imparts to Initial Slot Decoder, which induces $h'^S_t$ to fit $\Tilde{h}^{[L, t]}_0$:
\begin{equation}
    \mathcal{L}_{MSE}\overset{\Delta}=- \frac{1}{n}\sum^n_{t=1}(h'^S_t-\Tilde{h}^{[L, t]}_0)^2
\label{eq:l2loss}
\end{equation}
As a joint training model, the parameters of two tasks are optimized jointly. We use the NLLloss and BCEWithLogitsLoss for slot filling and multiple intent detection respectively. The total loss is:
\begin{equation}
    \mathcal{L}_{total}=\alpha\cdot\mathcal{L}_{MSE}+\beta\cdot\mathcal{L}_{NLL}+\lambda\cdot\mathcal{L}_{BCE}
\label{eq:totalloss}
\end{equation}
with three hyper-parameters $\alpha$, $\beta$, and $\lambda$ to balance them.

\renewcommand{\arraystretch}{1.05} 
\begin{table}[ht]
\centering
\fontsize{6.5}{6.5}\selectfont
\begin{threeparttable}
\caption{Slot filling and multiple intent detection results on two multi-intent datasets. ``$cat$" stands for concat and ``$sig$" stands for sigmoid.}
\label{tab:main_results}
\begin{tabular}{lcccccc}
\toprule
\multirow{3}{*}{Model}&
\multicolumn{3}{c}{ MixATIS}&\multicolumn{3}{c}{ MixSNIPS}\cr
\cmidrule(lr){2-4} \cmidrule(lr){5-7}
&Slot&Intent&Overall&Slot&Intent&Overall\cr
&(F1)&(Acc)&(Acc)&(F1)&(Acc)&(Acc)\cr
\midrule
Slot-Gated ($cat$) \cite{goo2018slot}& 87.7& 63.9& 35.5& 87.9& 94.6& 55.4\cr
Bi-Model ($cat$) \cite{Wang2018ABB}& 83.9& 70.3& 34.4& 90.7& 95.6& 63.4\cr
SF-ID ($cat$) \cite{Haihong2019ANB}& 87.4& 66.2& 34.9& 90.6& 95.0& 59.9\cr
Stack-Propagation ($cat$) \cite{Qin2019ASF}& 86.5& 76.2& 39.6& 93.7& 96.2& 72.4\cr
Stack-Propagation($sig$) \cite{Qin2019ASF}& 87.8& 72.1& 40.1& 94.2& 96.0& 72.9\cr
Joint Multiple ID-SF \cite{gangadharaiah-narayanaswamy-2019-joint}& 84.6& 73.4& 36.1& 90.6& 95.1& 62.9\cr
AGIF \cite{qin-etal-2020-agif}& 86.7& 74.4& 40.8& 94.2& 95.1& 74.2\cr
\midrule
SDJN& \textbf{88.2}&{\bf 77.1}&{\bf 44.6}& \textbf{94.4}&{\bf 96.5}&{\bf 75.7}\cr
\bottomrule
\end{tabular}
\end{threeparttable}
\end{table}

\renewcommand{\arraystretch}{1.} 
\begin{table*}[ht]
\centering
\fontsize{7.5}{7.5}\selectfont
\scalebox{0.95}{
\begin{threeparttable}
\caption{Ablation study on MixATIS. We try two types of distilled knowledge sources. One is soft targets: ``Soft" with the temperature setting ``$Temp$" under, and the other is hint: ``Hint", which means the output of the hidden layers from the teacher model. ``Implicit" represents that decoders only share the same encoder, while ``Explicit" means the output of one decoder will serve as auxiliary information for the next Decoder.}
\label{tab:ablation}
\begin{tabular}{cccccccccccc}
\toprule
\multicolumn{9}{c}{SDJN Components}&\multicolumn{3}{c}{ MixATIS}\cr
\cmidrule(lr){1-9} \cmidrule(lr){10-12} 
&Initial Slot& MIL Intent&Final Slot& Soft& Soft&\multirow{2}{*}{Hint}& \multirow{2}{*}{Implicit}& \multirow{2}{*}{Explicit}&\multirow{2}{*}{Slot (F1)}&\multirow{2}{*}{Intent (Acc)}&\multirow{2}{*}{Overall (Acc)}\cr
&Decoder& Decoder &Decoder& $Temp$=2 & $Temp$=4& & & &&&\cr
\midrule
(a)&\CheckmarkBold& \CheckmarkBold& & & & & \CheckmarkBold& & 86.5& 73.1& 36.7\cr
(b)&\CheckmarkBold& \CheckmarkBold& & & & & & \CheckmarkBold&88.0& 75.9& 42.0\cr
(c)&\CheckmarkBold& \CheckmarkBold& \CheckmarkBold& & & & & \CheckmarkBold& 88.1& 76.1& 43.0\cr
(d)&\CheckmarkBold& \CheckmarkBold& \CheckmarkBold& \CheckmarkBold& & & &\CheckmarkBold &87.5&{\bf 77.4}&43.5\cr
(e)&\CheckmarkBold& \CheckmarkBold& \CheckmarkBold& &\CheckmarkBold& & &\CheckmarkBold &{\bf 88.3}&76.3&43.1\cr
(f)&\CheckmarkBold& \CheckmarkBold& \CheckmarkBold&  & & \CheckmarkBold& & \CheckmarkBold&88.2&77.1&{\bf 44.6}\cr
\bottomrule
\end{tabular} 
\end{threeparttable} }
\end{table*}

\section{Experiments}
\label{sec:majhead}

\subsection{Datasets}
\label{ssec:Datasets}
We conduct our experiments on two public multi-intent NLU datasets.~\nocite{huang2021GhostBERT,hou2021DynaBERT} They are the cleaned version of MixATIS \cite{qin-etal-2020-agif} and MixSNIPS \cite{qin-etal-2020-agif}. MixATIS dataset is collected from ATIS dataset \cite{hemphill-etal-1990-atis} and MixSNIPS dataset is from SNIPS dataset \cite{Coucke2018SnipsVP}. Both of the multi-intent NLU datasets have the ratio of sentences with 1 $\sim$ 3 intents as [0.3, 0.5, 0.2]. In MixATIS dataset, there are 13162, 759, 828 utterances for training, validation, and testing. MixSNIPS dataset has 39776, 2198, 2199 utterances for training,validation, and testing.

\subsection{Experimental Setup}
\label{ssec: Experimental Setup}
We set the self-attentive encoder hidden units as 256, dropout rate as 0.4, threshold $t_i$ as 0.5 emperically with Adam optimizer for both datasets. For batch size, we set 16 and 64 for MixATIS and MixSNIPS. The hyper-parameters of loss are empirically set as $\alpha$: $\beta$: $\lambda$= 1: 0.7: 0.6 for MixATIS and  $\alpha$: $\beta$: $\lambda$= 1.25: 1: 1 for MixSNIPS. For all experiments, we select the model which performs the best on the dev set and evaluate it on the test set. We evaluate the performance of slot filling with F1 score, intent detection with macro F1 score and accuracy, and the NLU semantic frame parsing with overall accuracy that represents all metrics are correct in the utterance.

\subsection{Main Results}
\label{ssec:results}

The main results from the datasets are shown in Table \ref{tab:main_results}. There are seven baselines, including single-intent NLU and multiple-intent NLU. For single-intent NLU baselines handling multi-intent utterance, multiple intent labels are connected with ``$\#$" and treat as a single multi-intent label for comparison, named as $concat$ version. The $sigmoid$ version of Stack-Propagation predicts multi-intent labels directly by replacing softmax as sigmoid in the intent decoder with the binary cross-entropy loss. 

As we can see, our model outperforms all baselines on both datasets. For Slot (F1) score, our model shows strong performance and outperforms the best baseline, AGIF, 1.5\% and 0.2\% on MixATIS and MixSNIPS, showing the advantages of adopting fine-grained multiple intent information for slot filling. For Intent (Acc), it is easier for the $concat$ version of single-intent baselines to predict multiple intents because concatenation can greatly reduce the multi-intent search space \cite{qin-etal-2020-agif}. Though facing difficulty, our model yet outperforms the previous top score baseline, Stack-Propagation ($cat$), 0.9\% and 0.3\%  on MixATIS and MixSNIPS respectively, showing the effectiveness of detecting multiple intents with MIL and using slot information to guide intent prediction. For overall (Acc), the improvements are 3.8\% and 1.5\% on MixATIS and MixSNIPS. It indicates that SDJN model can better correlate the relation between multiple intents and slots and further improve the whole NLU semantic frame parsing. We attribute the results to the fact that our model benefits the NLU performance from the auxiliary loop and distillation method.

\subsection{Ablation Study}
\label{sec:Ablation}
\subsubsection{Effect of each Decoder}We analyze how three decoders work with an ablation study on the MixATIS dataset as illustrated in Table \ref{tab:ablation} with rows (a)(b)(c). The experiments are conducted by gradually adding each decoder and whether to adopt explicit interactions between slots and intents.

As shown in the results, with the slot information from the Initial Slot decoder, row (b) is able to outperform row (a) on every metric significantly. The increment of 5.3\% on overall accuracy shows the effectiveness of Initial Slot Decoder guiding multiple intent detection. We believe that the MIL Intent Decoder benefits a lot from the aligned token-level slot information. Comparing row (b) and row (c), with the Final Slot decoder adding on to row (c), the results again show improvements on every metric. This ablation study shows that each decoder contributes improvements and considering the cross-impact between slot filling and multiple intent detection brings better results.

\subsubsection{Effect of Distillation Method}To further examine the effectiveness of our distillation method, we show the ablation study in the bottom part of Table \ref{tab:ablation} with rows (c)(d)(e)(f). From Table \ref{tab:ablation}, all rows with distillation method (rows (d)(e)(f)) outperform row (c) from 0.1\% up to 1.6\% in overall accuracy. It shows the effectiveness of the proposed self-distillation method. Comparing row (d) and row (e), we find it interesting that while both using soft targets as a knowledge source, row (d) with $Temp$=2 shows better performance in multiple intent detection and overall accuracy while row (e) with $Temp$=4 shows better performance in slot filling. Comparing the distilled knowledge source, row (f) which uses hint as knowledge source outperforms rows (d)(e) that both use the soft target as a knowledge source 1.1\% and 1.5\% respectively in overall accuracy. We argue that the soft targets mainly rely on the output of the last layer of the decoder and fail to address the intermediate-level supervision which is important for representation learning. On the other hand, hint offers intermediate-level supervision and enables the model to learn better representation. Therefore, row (f) has a more balanced performance in slot filling and multiple intent detection, which further leads to a better result in overall accuracy.

\section{Conclusion}
\label{sec:page}
In this work, we propose a Self-distillation Joint NLU model by taking advantage of multi-task. In addition, we approach multiple intent detection as a weakly supervised problem with MIL. Experiments on two public multi-intent datasets show that our method achieves performance gains over strong baselines.


\bibliographystyle{IEEE.bst}
\bibliography{refs.bib}

\end{document}